\documentclass[graybox]{svmult}

\usepackage{amsmath}
\usepackage{mathptmx}       

\usepackage{helvet}         
\usepackage{courier}        
\usepackage{type1cm}        
\usepackage{makeidx}         
\usepackage{graphicx}        
\usepackage{multicol}        
\usepackage[bottom]{footmisc}

\newcommand\independent{\protect\mathpalette{\protect\independenT}{\perp}}
\def\independenT#1#2{\mathrel{\rlap{$#1#2$}\mkern2mu{#1#2}}}
\newcommand{\probP}{\text{I\kern-0.15em P}}

\makeindex             

\begin{document}

\title*{Improving Open-Domain Dialogue Evaluation with a Causal Inference Model}
\author{Cat P. Le, Luke Dai, Michael Johnston, Yang Liu, Marilyn Walker, Reza Ghanadan}
\institute{Cat P. Le \at Duke University, \email{cat.le@duke.edu}.\\ This work was done while Cat P. Le was a research intern at Amazon Alexa AI.
\and Luke Dai, Michael Johnston, Yang Liu, Marilyn Walker, Reza Ghanadan \at Amazon Alexa AI, \email{\{lukedai, mjohnstn, yangliud, wamarilk,ghanadan\}@amazon.com}}
%
%
\maketitle

\abstract{Effective evaluation methods remain a significant challenge for research on open-domain conversational dialogue systems. Explicit satisfaction ratings can be elicited from users, but users often do not provide ratings when asked, and those they give can be highly subjective. Post-hoc ratings by experts are an alternative, but these can be both expensive and complex to collect. Here, we explore the creation of automated methods for predicting both expert and user ratings of open-domain dialogues. We compare four different approaches. First, we train a baseline model using an end-to-end transformer to predict ratings directly from the raw dialogue text. The other three methods are variants of a two-stage approach in which we first extract interpretable features at the turn level that capture, among other aspects, user dialogue behaviors indicating contradiction, repetition, disinterest, compliments, or criticism. We project these features to the dialogue level and train a dialogue-level MLP regression model, a dialogue-level LSTM, and a novel causal inference model called counterfactual-LSTM (CF-LSTM) to predict ratings. The proposed CF-LSTM is a sequential model over turn-level features which predicts ratings using multiple regressors depending on hypotheses derived from the turn-level features. As a causal inference model, CF-LSTM aims to learn the underlying causes of a specific event, such as a low rating. We also bin the user ratings and perform classification experiments with all four models. In evaluation experiments on conversational data from the Alexa Prize SocialBot, we show that the CF-LSTM achieves the best performance for predicting dialogue ratings and classification.}


\section{Introduction}
\label{sec:intro}
Evaluation has always been a complex challenge for interactive dialogue systems. For task-oriented dialogues, frameworks such as Paradise~\cite{walker-etal-1997-paradise} model the relationship between user satisfaction, task completion, and cost factors such as dialogue length, word error rate, and dialogue behaviors. However, open-domain dialogue systems such as those built for the Alexa Prize SocialBot Grand Challenge~\cite{ram2018conversational,gabriel2020further}, where there is no clearly defined task, require new metrics and methods for evaluation that better reflect their affordances \cite{walker2021modeling,kim2020speech,ghazarian-etal-2019-better,ghazarian2022wrong,higashinaka2019improving,sinha-etal-2020-learning}.

In Alexa Prize SocialBot, user conversation ratings are elicited primarily for comparative evaluation of systems competing in 
the Grand Challenge.
Numerous other use cases for dialogue ratings include selecting training data and running A/B tests to evaluate new capabilities, features, and models. Manually collected user ratings are limited in that only a fraction of users of ``Alexa Let's Chat" leave ratings, and these ratings  can be highly subjective. An alternative is the post-hoc annotation of dialogues by trained annotators, which is complex, expensive, and thus not scalable. Consequently, a significant amount of dialogue data is unavailable for experimentation and modeling tasks.

To overcome this problem and increase the number of rated dialogues, we evaluate four methods for automatically predicting both user (128K) and expert (1.2K) ratings  of de-identified conversations collected from interactions with an Alexa SocialBot. 
First, we train a baseline method using a transformer-based model~\cite{devlin-etal-2019-bert} to predict dialogue ratings based on the raw input dialogue sequence. The other three methods are variants of a two-stage approach in which we use pre-trained models to extract interpretable features at the turn level.
These models include  classifiers we call ODES  (Open Domain Evaluation Signals) that detect user inputs such as  criticism, insults, compliments, requests to stop, and statements that indicate misunderstanding, repetition, or contradiction among other classes~\cite{walker2021modeling}. DialoGPT~\cite{zhang2020dialogpt} and DialogRPT~\cite{gao-etal-2020-dialogue} are applied to the dialogue sequence to provide measures of response relevance and specificity, and FED~\cite{mehri2020} is applied to each turn-pair to provide  metrics for the interestingness, engagingness, specificity, relevance, correctness, semantic appropriateness, understandability and fluency of responses. In addition to these derived measures, we also include ASR confidence and sentiment scores based on  processing the acoustics of user utterances. These features are projected to the dialogue level as averages, counts, initial, final, and penultimate values and used as input to our three models.
 
The first model trains an MLP regressor to predict dialogue ratings. The second model, called dialogue-level LSTM, uses an LSTM model on the sequence of turn-level features. The third approach models the rating using a shared sequential model over the turn-pair level vectors and introduces a causal structure into the prediction model. This proposed Counterfactual-LSTM (CF-LSTM) is a causal inference model that uses different models to predict the overall rating under different hypotheses derived from the ODES classifiers. In particular, a treatment (e.g., $0$ or $1$) is applied to each dialog based on the text classification. For instance, dialog with negative texts, such as critique, disinterest, and insult, will be assigned a treatment of $1$. On the other hand, dialogs without negative texts will be assigned a treatment of $0$. Next, a causal inference neural network is trained using the dialog features and their corresponding treatments to predict the customer's rating. This approach introduces the domain knowledge into the prediction model and helps further improve the performance. Multiple different regression functions are used based on the hypothesis. Lastly, we apply all four models  to classify dialogues into bins according to their ratings (e.g., binary and 5-class classification). Our contributions are:
\begin{enumerate}
    \item We identify and train classifiers to extract various types of dialogue features (e.g., sentiment, response relevance, and specificity) and use these features to predict dialogue ratings.
    \item We observe the special conditions of the open-domain dialogue evaluation problem and make three assumptions to apply causal inference to this problem.
    \item We introduce three approaches and propose a novel causal inference model, CF-LSTM, that is used to evaluate open-domain dialogues via mapping hidden dialogue features to the dialogue ratings.
\end{enumerate}

\section{Related Work}
\label{sec:related}
Much recent work on dialogue evaluation has focused on  the utterance level, measuring the goodness of a particular system response in the dialogue context. This type of evaluation can be either reference-based or referenceless. Reference-based approaches re-apply  metrics from tasks such as machine translation or summarization and compare a proposed system utterance to a reference response, 
which may be  from a pre-existing dialogue  \cite{gopalakrishnan2019topical}, or collected via crowd-sourcing  \cite{juraska2019viggo}. For instance, BLEU~\cite{papineni2002bleu} computes the n-gram precision of the system's response string when compared to a reference response, while METEOR~\cite{banerjee2005meteor} considers the stems and synonyms of the reference and ROUGE~\cite{lin2004rouge} uses n-gram recall instead of n-gram precision. More recent neural versions of these metrics, such as BERTScore and BLEURT, use word embeddings 
rather than the raw strings~\cite{zhang2019bertscore, scialom-etal-2021-questeval, sellam-etal-2020-bleurt}. However, it is well-known that these metrics have limited value for task-oriented dialogue and  even less utility when evaluating the highly varied responses typical of open-domain dialogues ~\cite{reiter_dale_1997,gatt2018survey,liu2016not}.

Work on referenceless  metrics aims to predict the quality of system response in the dialogue context without using reference utterances. These metrics
evaluate the proposed system utterance in terms of  coherence, interestingness, engagingness, fluency, specificity, relevance, empathy, or other measures. For example, 
GRADE measures coherence in terms of a graph-based representation of topic relatedness \cite{huang-etal-2020-grade}. 
DialogRPT  can be used to provide  metrics that
estimate the likelihood that
an utterance
generates many responses, directly or indirectly  \cite{gao-etal-2020-dialogue},   based on training with  133M utterance pairs from Reddit, consisting of utterances and their upvotes, downvotes, and number of responses.

Another idea  is to score a proposed system utterance by the probability that it elicits 
a particular user response type, such as disinterest or  criticism \cite{shalyminov2018neural, mehri2020unsupervised,walker2021modeling,ghazarian2022wrong}.
For example, the FED framework scores proposed utterances   by utilizing the DialoGPT LM probabilities for the subsequent user utterances such as ``That is interesting'' \cite{zhang2019dialogpt,mehri2020unsupervised}. Predictive models can also be generated from large dialogue corpora by using the following user utterance as weak supervision  \cite{ghazarian2022wrong,shalyminov2018neural}.


Other research aims to predict the overall quality of the dialogue by developing models for predicting a quality measure in terms of particular types of dialogue events. The most straightforward framework is end-to-end (E2E) approaches using the raw dialogue text as input to a transformer model~\cite{ghazarian-etal-2019-better, devlin-etal-2019-bert, ghazarian2020predictive}. This approach performs well with expert-rated data but  does not predict  user-assigned ratings well. We use this approach as our E2E baseline below.

Other work first derives several intermediate metrics at the turn or utterance level and then uses those to predict the dialogue quality, as we do here. The first framework like this was PARADISE, which included task completion as an input to the model, along with other metrics such as ASR error rates, system dialogue acts, or user requests to start over \cite{walker-etal-1997-paradise,walker2001date}. Recent work on PARADISE has used other features, such as user complaints and compliments, to apply it to open-domain dialogue \cite{walker2021modeling}. Many recent approaches work with embedding features, such as sentiment scores and utterance classification, before predicting the user rating~\cite{zhang2021dynaeval, ghazarian2020predictive}. Sentiment analysis~\cite{kim2020speech, ghazarian2022wrong} and dialogue features analysis~\cite{mehri2020, see2019makes, walker2021modeling} methods, for instance, show solid results for predicting user-assigned ratings.


Furthermore, many approaches to open-domain dialogue evaluation use large-scale corpora~\cite{charras2016comparing, duplessis2016purely, dodge2015evaluating, hori2017end, serban2016building}. Since the Alexa Prize challenge was introduced in recent years, several new controlled dialogue datasets have been created~\cite{zhang2018personalizing, rashkin2018towards, li2017dailydialog, gopalakrishnan2019topical}. However, evaluating the Alexa Prize SocialBot remains a challenging problem due to the nature of the collected dialogues. These dialogues are de-identified conversations between thousands of users and the Alexa SocialBot covering numerous topics and genres, with widely varying lengths (e.g., the SocialBot dialogues vary in length  from $3$ to $230$ turn-pairs). Additionally, the SocialBot is expected to have up-to-date knowledge about news, sports, movies, music, books, etc. Hence, the evaluation of the Alexa SocialBot requires a  dynamic structure that takes into account domain knowledge. 

In this paper, we introduce a novel causal inference structure to open-domain dialogue evaluation  to improve the performance of the prediction model for the Alexa SocialBot.
Causal inference~\cite{rubin1978bayesian, rosenbaum1983central} has been extensively investigated in the neural network literature~\cite{feder2021causal, shalit2017estimating, alaa2017bayesian, yoon2018ganite, aloui2022causal} because of the black box nature of neural models. Notably, the TARNet model~\cite{shalit2017estimating}, a causal inference neural network, has been applied to causal inference for various regression and classification problems. It can be applied to many machine learning models, such as image generative models~\cite{yoon2018ganite} and Gaussian processes~\cite{alaa2017bayesian}. However, we have yet to be aware of previous work that uses causal inference for open-domain dialogue evaluation. Here we propose a novel LSTM causal structure for predicting the quality of open-domain dialogue to make the evaluation model more robust.

\section{Proposed Methods}
\label{sec:method}
\begin{figure}[th!]
\centering
\centerline{\includegraphics[width=\textwidth]{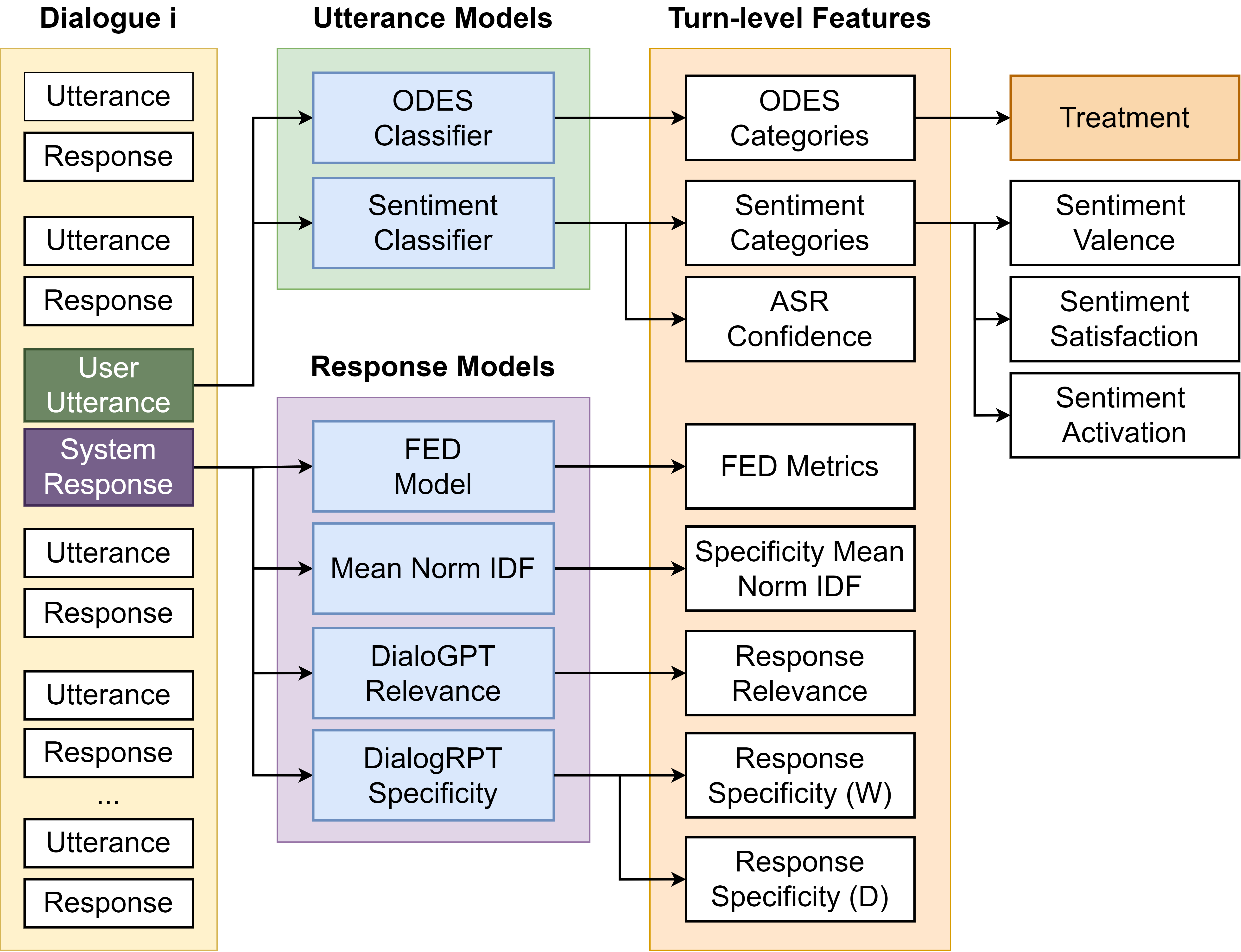}}
\caption{The dialogue turn-level feature extractors consist of $2$ utterance models (i.e., ODES classifier, sentiment classifier) and $4$ response models (i.e., FED model, Mean Norm IDF, DialoGPT Relevance, and DialogRPT Specificity). The treatment assignments for our causal model are extracted using the ODES classifiers.}
\label{fig:extractor}
\end{figure}

We propose three approaches to tackle  open-domain dialogue evaluation. These approaches convert each turn pair in the dialogue into a vector representation, then process the vectors to predict the dialogue level rating. Our proposed methods  consist of two main components.
\begin{enumerate}
    \item \textbf{Dialogue Turn-level Feature Extractors}. These extractors are applied to extract meaningful hidden features from the dialogue as in Figure~\ref{fig:extractor}. Subsequently, these turn-level features are converted to the dialogue-level features that are fed into the prediction models.
    \item \textbf{Prediction Models}. We utilize an end-to-end transformer on the raw dialogue text to create a baseline. We then compare this baseline to  three proposed models for predicting both user and expert ratings from the hidden dialogue-level features:
    \begin{itemize}
        \item Dialogue-level MLP (Section~\ref{other_model})
        \item Dialogue-level LSTM (Section~\ref{other_model})
        \item Counterfactual LSTM (Section~\ref{model})
    \end{itemize}
\end{enumerate}
The dialogue-level MLP uses an MLP regressor to project the turn features to the user ratings, while the dialogue-level LSTM utilizes a standard LSTM to predict ratings from the sequence of turn-level features. The counterfactual LSTM (CF-LSTM) is a novel causal inference model. It consists of the LSTM layers augmented with the causal analysis model. This causal model consists of multiple MLP regressors, each trained to predict the ratings based on a specific hypothesis.

\subsection{Dialogue Turn-level Feature Extractors}
\label{feature_extractor}
 
To extract  features for each utterance-response pair, we apply $6$ models (i.e., $2$ models for the utterances and $4$ models for the responses), as shown in Figure~\ref{fig:extractor}. We apply the Open-Dialogue Evaluation Signals (ODES) Classifier and an Audio-based Sentiment Classifier for the utterance data. The ODES classifier, which is based on the T5 architecture~\cite{raffel2020exploring}, was trained on 8K annotated utterances to categorize them into the classes shown in Table~\ref{table_example}:  (i) user disinterest, (ii) user critique, (iii) user not understand, (iv) user requests topic switch, (v) user obscenity, (vi) user rejects topic switch, (vii) user requests repeat, (viii) user requests to stop, (ix) user insult, (x) user compliment, (xi) user calls out repetition, (xii) user calls out contradiction, (xiii) system not understand, and (xiv) others. This T5 classifier achieves above $95\%$ accuracy on the testing corpora. Next, the ODES categories assign the treatment $T$ for the CF-LSTM. 

\begin{table}[h!t]
\caption{\label{table_example} Examples of dialogues for each  ODES category in the Alexa SocialBot dataset.}
\begin{center}
\begin{tabular}{c|c|c|p{2.7in}}
\noalign{\smallskip}\svhline\noalign{\smallskip}
{\bf Class}  & {\bf ODES Name}  & {\bf Counts} & {\bf Example}\\
\hline\hline
i   & User              & 10,938    &  \textbf{SocialBot}: Did you know that otters sleep holding hands? \\ 
    & disinterest       &           &  \textbf{User}: I really couldn't care less. \\
\hline
ii  & User              & 12,290      &  \textbf{SocialBot}: Who do you think will win Superbowl 2019?\\
    & critique          &           & \textbf{User}: You are really a very stupid bot.\\
\hline
iii & User not          & 12,399    &  \textbf{SocialBot}: What is your favorite genre of video game? \\ 
    & understand        &           &  \textbf{User}: I don't know what genre means. \\
\hline
iv  & User requests     & 21,278    &  \textbf{SocialBot}: Who is your favorite Batman actor?\\ 
    & topic switch      &           &  \textbf{User}: Can we talk about something else?\\
\hline
v   & User              & 66,532    &  \textbf{SocialBot}:  Do you have a favorite movie?\\ 
    & obscenity         &           &  \textbf{User}: B*** me.\\
\hline
vi  & User rejects      & 4,389     &  \textbf{SocialBot}: I love country music too. But I was wondering,\\
    & topic switch      &              & do you have a favorite sport?  \\
    &                   &              &  \textbf{User}: No, I want to keep talking about music.\\
\hline
vii & User requests         & 26,779    &  \textbf{SocialBot}: Would you rather climb the Eiffel Tower or the \\
    & to repeat &           & Empire State Building? \\
    &           &           &  \textbf{User}: Could you say that again?\\
\hline
viii & User requests        & 78,504    &  \textbf{SocialBot}: I just love talking about music. What is your \\
    & to stop   &           & favorite kind of music?\\ 
    &           &           &  \textbf{User}: Please stop I need to go to bed.\\
\hline
ix  & User insult             &12,173     &  \textbf{SocialBot}: Wow, Taylor Swift has 97 albums. That's a lot!\\ 
    &             &           &  \textbf{User}: You are so full of sh**.\\
\hline
x   & User              & 57,539    &  \textbf{SocialBot}: Did you know LeBron James Jr has a college \\
    & compliment        &           & football scholarship, and he's only 12 years old?\\ 
    &        &           &  \textbf{User}: That's really interesting. \\
\hline
xi  & User calls out    & 15,052    &  \textbf{SocialBot}: What's your favourite football team?\\ 
    & repetition        &           &  \textbf{User}: You already asked me that question twice.\\
\hline
xii & User calls out    & 3,147     &  \textbf{SocialBot}: I don't have any pets.\\ 
    & contradiction     &           &  \textbf{User}: You just said you had a cat.\\
\hline
xiii& System not        & 12,534    &  \textbf{User}: Can we talk about Elle King?\\
    & understand        &           &   \textbf{SocialBot}: I like BB King too.\\ 
    &                   &           &  \textbf{User}: That's not what I said.\\
\hline
xiv& Others            & 2,320,515 &  \textbf{SocialBot}: Do you like K-Pop music?\\ 
    &                   &           &  \textbf{User}: Yes, I often listen to Blackpink and BTS.\\
\hline
\end{tabular}
\end{center}
\end{table}

We also extract $3$ different types of utterance sentiment features: sentiment valence, satisfaction, and activation, which are highly correlated with user ratings ~\cite{kim2020speech, ghazarian2022wrong}. 

To generate scores for the responses, we apply the FED model~\cite{mehri2020}, the DialoGPT model~\cite{zhang2019dialogpt}  for response relevance, the DialogRPT model\cite{gao-etal-2020-dialogue}  for response specificity width and depth, and Mean Norm IDF as another measure of response specificity~\cite{mehri2020, zhang2019dialogpt}. The concatenation of these extracted features forms the representation of the utterance-response pair, called turn-level features. These features are normalized to their Z-scores before being converted into dialogue-level features to train the prediction models.

\subsection{Dialogue-level MLP and LSTM}
\label{other_model}

The dialogue-level MLP is the two-level approach that uses the MLP~\cite{10.5555/1639537.1639542} regression to project the turn pair features to the ratings. Here, we consider the dialogue features that are an aggregation of the turn-level features by choosing a few fundamental values, such as average, initial, final, and penultimate values of the sequence of turn-level features. In this case, the dialogue features (i.e., input of the Dialogue-level MLP) have the same length and can be fed into the MLP regressor for prediction.

The dialogue-level LSTM is a method that uses a standard LSTM~\cite{10.1162/neco.1997.9.8.1735} on the dialogue-level features (i.e., a sequence of turn-level features) as in Table~\ref{feature_extractor}.  This model is similar to the one used for speech sentiment analysis~\cite{kim2020speech}, in which the LSTM model is used to map the sequence of turn-level sentiment features (i.e., sentiment valence, sentiment satisfaction, sentiment activation sentiment scores) to the ratings. Here, the input of the model is the concatenated turn-level features. Since this model is based on LSTM layers, it can handle the varied-length input (e.g., dialogue-level feature vectors). 

\subsection{Counterfactual LSTM}
\label{model}

\begin{figure}[b]
\sidecaption
\includegraphics[width=0.45\textwidth]{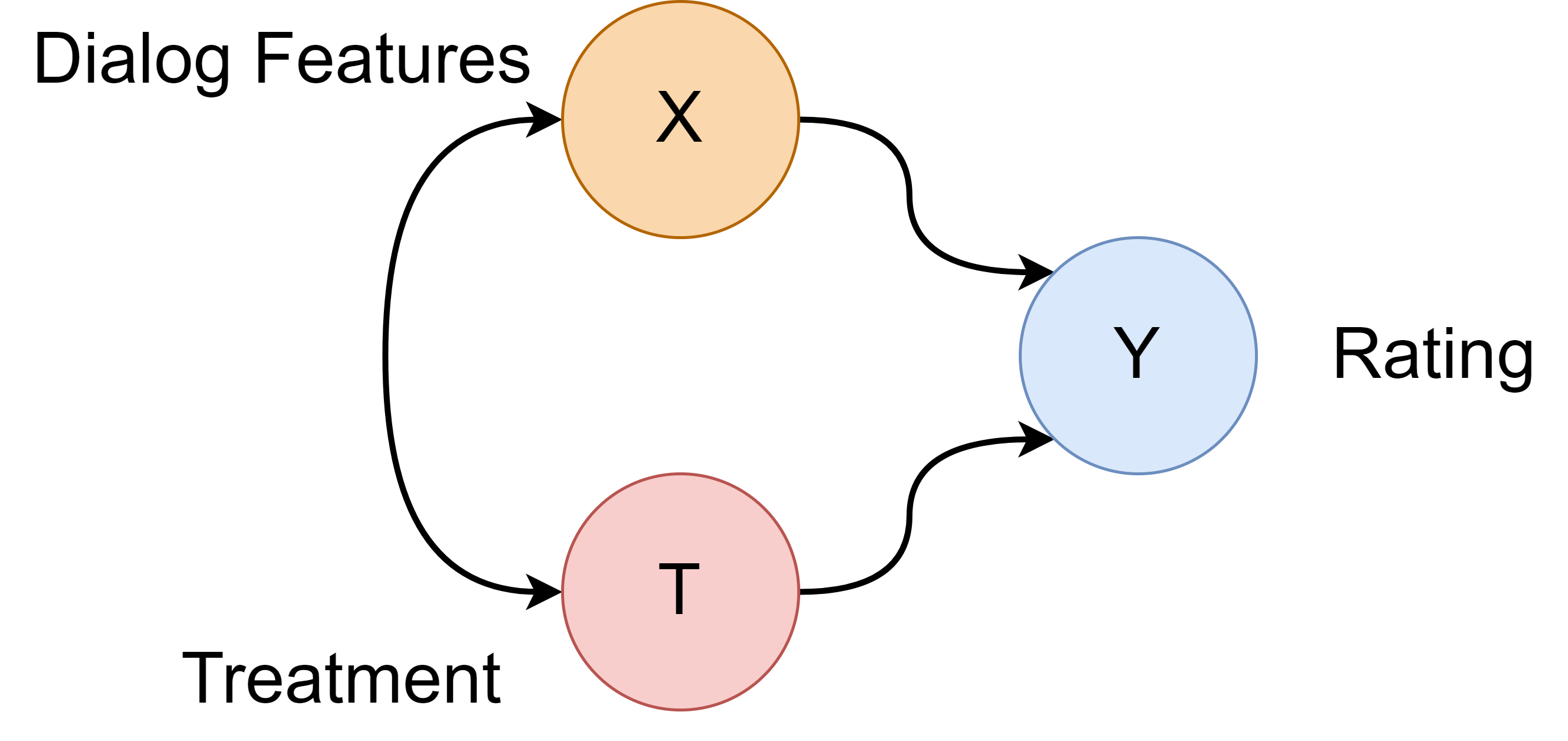}
\caption{The causal relationship between outcomes, treatments, and input features. The outcome $Y$ (e.g., customer ratings) is a function of the dialogue features $X$, and the treatments $T$.}
\label{fig:assumption}
\end{figure}

In recent natural language processing (NLP) tasks, the assumption that training and test data are identically distributed is often not satisfied. Bias and variance from the data are often introduced into the standard NLP approaches (e.g., non-causal models), which makes the training procedure difficult, and leads to poor results in real-world test data. We propose using a causal structure containing domain knowledge to overcome this issue. This structure helps introduce inductive biases and leads to more robust predictors~\cite{feder2021causal}. To apply causal inference to our problem, three common conditions must be satisfied: Ignorability, Positivity, and Consistency~\cite{feder2021causal}. 

\begin{itemize}
    \item \textbf{Ignorability}. In causal inference and statistics, the ignorability requires that the data collection method be independent of the treatment assignment. In other words, the outcomes (i.e., factual and counterfactual) do not depend on the treatment assignment.
    \begin{equation}
        T \independent Y(T=a), \quad \forall a \in \{0, 1\}
    \end{equation}
    \item \textbf{Positivity}. The probability of a sample being assigned the treatment $T=a$ for $a \in \{0, 1\}$ is bounded between 0 and 1.
     \begin{equation}
        0 < \probP(T=a \big| X=x) < 1, \quad \forall a \in \{0, 1\}
    \end{equation}
    \item \textbf{Consistency}. A sample under a treatment $T=a$, for $a \in \{0, 1\}$ has an unique outcome. The outcome of a sample under treatment $T=a$ is identical to the outcome if that sample is again assigned to $T=a$.
     \begin{equation}
        T=a \iff Y(a) = Y, \quad \forall a \in \{0, 1\}
    \end{equation}
\end{itemize} 

Here, the positivity and consistency conditions are satisfied. To satisfy the ignorability condition, we must assume that the extracted dialogue features, described in Section~\ref{feature_extractor}, capture everything relevant about the dialogue to predict the ratings. In other words, we assume that all confounding variables $X$ are observed and that there is no hidden confounder. This property, called \textit{conditional ignorability}, is described as follows:
\begin{equation}
    \big(T \independent (Y(a)\big) \big| X, \quad \forall a \in \{0, 1\}
\end{equation}

We attempt to satisfy the ignorability condition by using many features in our models that were proposed in previous work, as well as introducing our own novel ODES classifiers, but we acknowledge that future work may derive additional
features from the dialogue events, such as dialogue acts or sub-dialogue level features \cite{walker2001date,prasad2002training}.


\begin{figure}[t]
\centering
\centerline{\includegraphics[width=\textwidth]{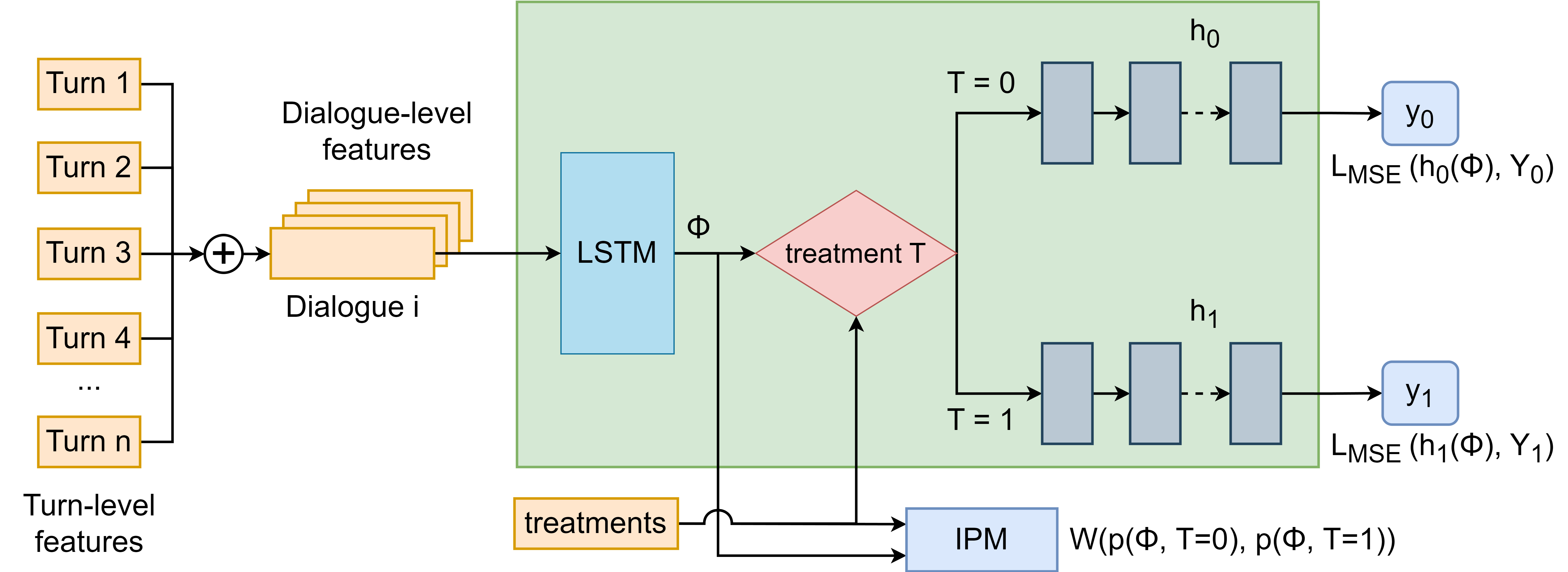}}
\caption{The diagram of the proposed counterfactual LSTM model. The model consists of the LSTM layers and the causal structure $h_0, h_1$. It learns to map the dialogue-level features to the user ratings based on the treatment $T$.}
\label{fig:cflstm}
\end{figure}

Next, we propose a novel causal inference model called the Counterfactual LSTM, which maps the dialogue-level features to the ratings based on a specific hypothesis. This paper considers a simple case with only $2$ hypotheses. 
First, the turn-level features are stacked into a dialogue-level feature vector for each dialogue. Since each dialogue has a different number of turn pairs, the dialogue-level features are varied-length vectors. These dialogue-level features represent the dialogue textual data fed into the CF-LSTM model to predict the dialogue ratings. As shown in Figure~\ref{fig:cflstm}, the proposed CF-LSTM model consists of the LSTM layers and $2$ individual MLP regressors, each used to train on a specific hypothesis or treatment. The inputs of the LSTM layers are the dialogue-level feature vectors, which are sequences of turn-level features. Since the LSTM can take varied-length inputs, this property also holds for CF-LSTM and can take varied-length dialogue-level feature vectors. Next, the output features $\Phi$ of the LSTM are fed to the specific MLP ($h_0$ or $h_1$) based on the hypothesis and trained to predict dialogue ratings. 


\subsubsection{Definition of Treatments}
\label{treatment}
We utilize the extracted ODES features from Section~\ref{feature_extractor} as the treatment groups. Let $T$ denote the treatment assignment. The treatment $T=0$ is the hypothesis group whose dialogues consist of only the ODES category (xiv), i.e., other. The treatment $T=1$ is the hypothesis group whose dialogues consist of one or more ODES categories (i) to (xii). Since the CF-LSTM has a dedicated MLP for each dialogue hypothesis, it can predict dialogue ratings for similar dialogue with a different hypothesis. For instance, two dialogues with different dialogue ratings have the same negative sentiment features. One dialogue is categorized into ODES group (i), i.e., user disinterest, and is assigned $T=1$.
On the other hand, the sentiment of the other dialogue is classified as unfavorable, and the user appears to be satisfied with the conversation, e.g., $T=0$. Non-causal regressors can quickly group these dialogues since they contain the same sentiment features. Still, the CF-LSTM with a causal structure can identify the difference between these two dialogues and better predict the rating. 

\subsubsection{Loss Function of CF-LSTM}
Here, we define the loss function for our proposed CF-LSTM model. First, most dialogues with the treatment $T=1$ have relatively low ratings. For example, the dialogues in group (viii), i.e., user insult, will often receive ratings of $2$ or less. Additionally, the ratings are given non-uniformly among different groups of users. These are the source of bias and variance in the dataset, and a large proportion of  users choose not to leave a rating. In order to avoid introducing variance and bias into the model, the Integral probability metric (IPM) is established in ~\cite{shalit2017estimating}. This metric is used as the upper bound for this source of variance. The IPM measures the distance between two distributions: $p(\Phi, T=0), p(\Phi, T=1)$. In this paper, we use Wasserstein distance~\cite{Villani2009} as the IPM measurement in our CF-LSTM model.
Let $Y_0$ and $Y_1$ denote the ground truth ratings for the treatment $T=0$ and $T=1$, respectively. The loss function for the CF-LSTM model is defined as follows:
\begin{equation}
\begin{aligned}
    L &= L_{MSE}\big(h_0(\Phi), Y_0\big|T=0\big) + L_{MSE}\big(h_1(\Phi), Y_1\big|T=1\big) \\
    &+ W\big(p(\Phi, T=0), p(\Phi, T=1)\big)
\end{aligned}
\end{equation}

where $L$ denotes the loss function of the CF-LSTM model, $L_{MSE}$ is the mean square error (MSE) loss, $W$ is the Wasserstein distance, $h_0()$ and $h_1()$ are the outputs of the MLP regressors $h_0$ and $h_1$, respectively. This loss function consists of two main parts, the MSE loss, and the IPM loss. The MSE loss is the error between the predicted and ground truth ratings under every hypothesis (e.g., $T=0$ and $T=1$). The IPM loss is defined as the Wasserstein distance between $2$ distributions of ratings described above. 

\subsubsection{Data Generation for Classification Tasks}
\label{data_generation}
\begin{figure}[t]
\centering
\includegraphics[width=.7\textwidth]{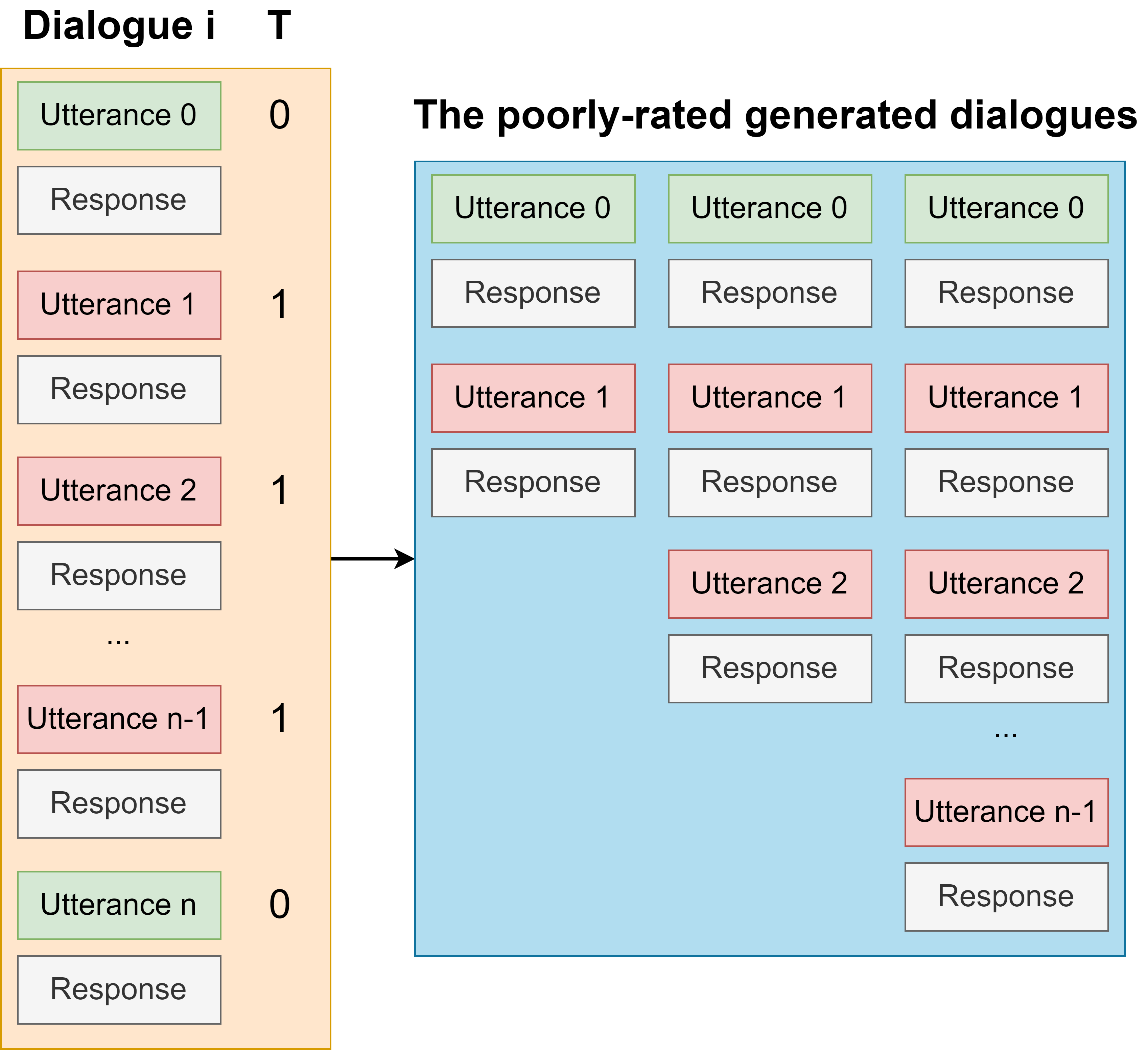}
\caption{The procedure of generating new dialogues by masking utterances. Let dialogue $i$ be the low-rated dialogue with $n+1$ utterances. Each utterance is given the treatment of $1$ and $0$. At turn $m$, if the treatment for utterance $m$ is $1$, we define a new poorly-rated dialogue with the utterances 0 to $m$ and masking the remaining utterances, i.e., utterances $(m+1)$ to $n$.}
\label{fig:mask}
\end{figure}

We consider the classification task of detecting low-quality dialogues at run-time. The model needs to be trained with sufficient data samples to dynamically identify the low-rated dialogues with high accuracy. It is often not satisfied in practice since most collected data samples have good ratings. As a result, we want to generate more dialogues with low ratings so that the model can be trained to identify these types of data quickly. It is also essential since it determines whether the model can be applied to evaluate dialogues at run-time. This online-evaluation mechanism can also improve the overall user experience since the system could navigate to a new topic if it realizes that the current conversation is rated poorly. Here, we first select the poorly-rated dialogue $i$ and augment these data by masking utterances. We can assign a treatment to each utterance in the dialogue from the extracted turn-level features. At utterance $m$, if the treatment is $1$, we define a new dialogue, labeled as $0$ (i.e., poorly rated dialogue), that consists of the utterance $0$ to the utterance $m$. In other words, we mask all of the utterances after $m$ from the original dialogue, as illustrated in Figure~\ref{fig:mask}. Lastly, this newly generated data is used to fine-tune the classification model. The resulting model can be applied online to dynamically improve the user experience at run-time or used as a data selection mechanism that identifies high-quality dialogues from low-quality dialogues.
 
\subsubsection{Transfer Learning with CF-LSTM}
As described above, the CF-LSTM is designed to predict the rating based on a fixed number of hypotheses (e.g., $2$) in the dataset. However, CF-LSTM can handle the increase or decrease in the number of hypotheses using transfer learning~\cite{9766163, le2021atask, le2021task, le2021improved, 9054118}. In other words, we can transfer the prior knowledge from the pre-trained CF-LSTM to a new causal NLP task using transfer learning techniques~\cite{9766163, le2021task}. For example, the new data samples arrive with added treatment types. Here, we modify the trained model by adding new MLP regressors (e.g., representing different hypotheses) to the existing structure. We initialize the weights randomly for the added regressors while freezing the other weights of the pre-trained model. Lastly, we update the entire model by fine-tuning it using the new data samples. In the next section, we discuss the experimental results of our proposed methods when applied to our real-world dataset for both regression and classification problems.

\section{Experimental Results}
\label{sec:result}
In the following experiments, we apply our proposed methods to a dataset of dialogues with user ratings collected from users interacting with the Alexa SocialBot~\cite{ram2018conversational}. This dataset consists of 128K de-identified conversational dialogues. On average, each dialogue has 20 utterances, and the data is limited to dialogue consisting of at least three turn-pairs. For each utterance in a dialogue, we assign treatments based on the ODES categories (Please see Section~\ref{treatment}). Table~\ref{table_example} indicates the number of utterances of each ODES category in the Alexa SocialBot dataset. When considering this dataset, $70\%$ of dialogues are assigned $T=0$, and $30\%$ of dialogues are assigned $T=1$. Next, we compare our methods to the baseline approach, the End-to-end (E2E) Transformer. It directly uses a transformer-based model~\cite{devlin-etal-2019-bert} to map the dialogue textual data into the user ratings.


First, we consider the regression problem of predicting the user ratings for the given dialogue. Let L1d denote the daily average prediction and L7d denote the 7-day rolling daily average prediction. Table~\ref{table1} shows the three methods' performances when compared with the E2E baseline in terms of the Pearson correlation between the predicted ratings and the ground truth. As shown in Table~\ref{table1}, CF-LSTM achieves higher correlations in the individual prediction, the daily average prediction, and the 7-day rolling daily average prediction. It also shows a performance boost compared to the Dialogue-level MLP approach, which suffers from information loss when converting turn-level to dialogue-level features.

\begin{table}[h!t]
\caption{\label{table1}The comparisons between open-dialogue evaluation methods for the regression problem in terms of Pearson correlation (e.g., individual, L1d, L7d predictions) and the classification problems (e.g., binary, 5-class) in terms of prediction accuracy, on SocialBot conversations}
\begin{center}
\begin{sc}
\begin{tabular}{lccccc}
\hline\noalign{\smallskip}
\multicolumn{1}{l}{}  &\multicolumn{1}{c}{\textbf{Individual}} &\multicolumn{1}{c}{\textbf{L1d$^\star$}} &\multicolumn{1}{c}{\textbf{L7d$^\dagger$}} &\multicolumn{1}{c}{\textbf{Binary}} &\multicolumn{1}{c}{\textbf{5-class}}\\
\multicolumn{1}{l}{\textbf{Methods}}  &\multicolumn{1}{c}{\textbf{Prediction}} &\multicolumn{1}{c}{\textbf{Prediction}} &\multicolumn{1}{c}{\textbf{Prediction}} &\multicolumn{1}{c}{\textbf{Class.}} &\multicolumn{1}{c}{\textbf{Class.}}\\
\noalign{\smallskip}\svhline\noalign{\smallskip}
E2E Transformer                 & 0.22  & 0.30  & 0.47  & 54.1\%  & 32.8\%\\ 
Dialogue-level LSTM             & 0.30  & 0.41  & 0.59  & 64.6\%  & 43.5\%\\ 
Dialogue-level MLP              & 0.31  & 0.40  & 0.66  & 62.5\%  & 46.1\%\\  
CF-LSTM                         & \bf 0.34  & \bf 0.46  & \bf 0.68  & \bf 67.8\%  & \bf 48.2\%\\ 
\noalign{\smallskip}\hline\noalign{\smallskip}
\end{tabular}
\footnotesize{\\$^\star$ daily average prediction, $^\dagger$ 7-day rolling average prediction}
\end{sc}
\end{center}
\end{table}

In addition, to understand the impact of assigning the treatment (i.e., ODES categories) to the dialogues, we consider the situation where the treatments of the collected dialogues are inverted (e.g., treatment $T=0$ becomes $T=1$, and vice versa). For the same dialogue feature vector, its treatment is inverted. This setting allows us to investigate the impact of the ODES Classifier when assigning the incorrect treatments and whether the models can overcome this misinformation. We feed the modified data samples to the well-trained CF-LSTM capable of predicting the counterfactual outcomes for these dialogues. Table~\ref{table3} shows the correlations between the predicted counterfactual outcomes and the original ground truth. Here, we want to investigate whether the model can predict correctly even if the treatment assignment is inverted (e.g., the ODES classifier malfunctions). As shown in Table~\ref{table3}, the individual prediction and daily average prediction (L1d) of the inverted treatments scenario still have a relatively high correlation with the original ground truth (i.e., when the treatments are not inverted). The mean square error between the factual outcomes $Y_1$ and the counterfactual outcomes $Y_0$ is $0.3034$. The average treatment effect (ATE) is defined as follows for our problem:
$$ATE = E[Y_{1i}-Y_{0i}]= -0.7809$$
The ATE shows that, on average, a dialogue being assigned treatment $T=1$ (e.g., consists of disinterest, insult) will have a lower user rating than the dialogue with $T=1$ by $0.7809$.

\begin{table}[t]
\sidecaption
\begin{sc}
\begin{tabular}{lcc}
\hline\noalign{\smallskip}
\multicolumn{1}{l}{\textbf{Pearson}}  &\multicolumn{1}{c}{\textbf{Original}} &\multicolumn{1}{c}{\textbf{Inverted}}\\
\multicolumn{1}{l}{\textbf{Correlation}}  &\multicolumn{1}{c}{\textbf{Treatments}} &\multicolumn{1}{c}{\textbf{Treatments}}\\
\noalign{\smallskip}\svhline\noalign{\smallskip}
Individual Prediction       & 0.34  & 0.26\\ 
L1d Prediction              & 0.46  & 0.35\\ 
L7d Prediction              & 0.68  & 0.52\\ 
\noalign{\smallskip}\hline\noalign{\smallskip}
\multicolumn{3}{l}{$^\star$ daily average, $^\dagger$ 7-day rolling average}
\end{tabular}
\end{sc}
\caption{\label{table3}The correlations of CF-LSTM when the treatment assignments for all dialogues are inverted.}
\end{table}

We randomly select a set of $1213$ dialogues and ask an expert to rate the conversational quality. Let $R_{expert}$ be the rating given by the expert, and $R_{user}$ be the ratings given by users. The expert ratings have a low Pearson correlation with the user ratings, demonstrating the challenge in predicting user ratings in open-domain dialogue since they are only loosely correlated with an expert evaluation of dialogue quality.
$$
\rho = corr(R_{expert}, R_{user}) = 0.2056
$$

Next, we use the CF-LSTM and the Dialogue-level MLP to predict the expert's rating in the given dialogues. Table~\ref{table4} indicates the correlations obtained from CF-LSTM and Dialogue-level MLP on expert-rated dialogues from Socialbot conversations. Due to the simplicity of the network architecture, the Dialogue-level MLP achieves competitive results in individual and daily average predictions. However, it performs poorly in 7-day rolling average predictions because of the sparsity across time of the small dataset of $1213$ dialogues. On the other hand, CF-LSTM performs well on all types of predictions and shows flexibility in training for both large and small datasets.

\begin{table}[t]
\sidecaption
\begin{sc}
\begin{tabular}{lcc}
\hline\noalign{\smallskip}
\multicolumn{1}{l}{\textbf{Pearson}}  &\multicolumn{1}{c}{\textbf{Dialogue-}} &\multicolumn{1}{c}{\textbf{CF-LSTM}}\\
\multicolumn{1}{l}{\textbf{Correlation}}  &\multicolumn{1}{c}{\textbf{level MLP}} &\multicolumn{1}{c}{\textbf{(ours)}}\\
\noalign{\smallskip}\svhline\noalign{\smallskip}
Individual Prediction       & 0.35  & 0.26\\ 
L1d Prediction              & 0.42  & 0.36\\ 
L7d Prediction              & 0.12  & 0.51\\ 
\noalign{\smallskip}\hline\noalign{\smallskip}
\multicolumn{3}{l}{$^\star$ daily average, $^\dagger$ 7-day rolling average}
\end{tabular}
\end{sc}
\caption{\label{table4}The correlations of prediction models on $1213$ expert-rated dialogues from Socialbot conversations.}
\end{table}

Lastly, we consider the problem of classifying the dialogues into 2-class (i.e., binary) and 5-class classification. For binary classification, we label any dialogue with a rating less than $3$ as 0 (e.g., low-rated dialogues). Hence, the remaining dialogues with ratings greater or equal to $3$ are labeled as $1$ (i.e., highly-rated dialogues). For the 5-class classification, we round up or down (i.e., round half-up) the ratings into five groups. For example, a dialogue with a rating of $4.5$ is classified into group $5$, and a dialogue with a rating of $3.4$ is classified into group $3$. In this dataset, we observe that most dialogues are often rated above $3$ out of $5$. As a result, we apply the data generation method described in Section~\ref{data_generation} to generate more poorly-rated dialogues. This newly generated data is used to fine-tune the binary classification model. The resulting model can be applied online to improve the user experience at run-time. Table~\ref{table1} indicates the performance of our model and other methods, in terms of classification accuracy, in binary and 5-class classification problems. Our proposed method CF-LSTM outperforms other approaches in both binary classification and 5-class classification.


\section{Conclusion and Future Work}
\label{sec:conclusion}
We propose a novel causal inference structure for open-domain dialogue evaluation. This structure utilizes turn-level features, such as sentiment analysis, textual categories, response relevance, and response specificity, and uses them to predict user ratings. Our method performs competitively in an evaluation with rated conversations with the Alexa SocialBot, in regression and classification problems. As a causal inference model, this CF-LSTM model is more robust in learning complex representations and capable of predicting the ratings for the dialogues under different hypotheses. In addition to offline uses, this model can also be applied at run-time to dynamically identify low-quality dialogues and trigger the introduction of different topics to help improve the user experience. This paper considers only $2$ treatments, which are all the ODES categories, versus others. In future works, we can further boost the flexibility of our model by increasing the number of treatments corresponding to the different ODES groups. Additionally, other essential dialogue features (e.g., text and sentiment embeddings) can be included in the input features to help improve the overall performance of the prediction model.

\newpage
\bibliographystyle{abbrv}
\bibliography{refs,athena}

\end{document}